\documentclass[10pt,twocolumn,letterpaper]{article}

\usepackage{cvpr} 
\usepackage{times}
\usepackage{epsfig}
\usepackage{graphicx}
\usepackage{amsmath}
\usepackage{amssymb}
\usepackage{cite}
\usepackage{kotex}
\usepackage{booktabs}
\usepackage[normalem]{ulem}
\usepackage{tikz}
\usepackage{xcolor}
\usepackage{stfloats}
\usepackage{placeins}
\usepackage{balance} 
\usepackage[none]{hyphenat}
 
\usepackage[pagebackref=true,breaklinks=true,
            colorlinks=true, linkcolor=blue, citecolor=blue, urlcolor=blue,
            bookmarks=false]{hyperref}

\begin{document}

\title{LLM-Assisted Cheating Detection in Korean Language via Keystrokes}

\author{Dong Hyun Roh \\
Bucknell University, USA\\
{\tt\small dhr014@bucknell.edu}
\and
Rajesh Kumar\\
Bucknell University, USA\\
{\tt\small rajesh.kumar@bucknell.edu}
\and
An Ngo\\
Bucknell University, USA\\
{\tt\small axn001@bucknell.edu}
}

\maketitle

\thispagestyle{empty}

\begin{abstract}
This paper presents a keystroke-based framework for detecting LLM-assisted cheating in Korean, addressing key gaps in prior research regarding language coverage, cognitive context, and the granularity of LLM involvement. Our proposed dataset includes $69$ participants who completed writing tasks under three conditions: Bona fide writing, paraphrasing ChatGPT responses, and transcribing ChatGPT responses. Each task spans six cognitive processes defined in Bloom's Taxonomy (remember, understand, apply, analyze, evaluate, and create). We extract interpretable temporal and rhythmic features and evaluate multiple classifiers under both Cognition-Aware and Cognition-Unaware settings. Temporal features perform well under Cognition-Aware evaluation scenarios, while rhythmic features generalize better under cross-cognition scenarios. Moreover, detecting bona fide and transcribed responses was easier than paraphrased ones for both the proposed models and human evaluators, with the models significantly outperforming the humans. Our findings affirm that keystroke dynamics facilitate reliable detection of LLM-assisted writing across varying cognitive demands and writing strategies, including paraphrasing and transcribing LLM-generated responses.\footnote{This paper has been accepted for publication at IEEE-IJCB 2025.} 
 
\end{abstract}

\section{Introduction}
\label{sec:introduction}
The rapid development and broad adoption of Large Language Models (LLMs) have raised serious concerns about academic integrity \cite{crossley2024,kundu2024,hoq2024,TypeNetNotInterpretableETS,plank2016, agarwal2022}. Traditional plagiarism and cheating detection tools, which rely on text comparison between submitted work and existing sources, often fail to detect more sophisticated forms of cheating, such as paraphrasing or transcription of LLM-generated content \cite{crossley2024,kundu2024,hoq2024, TypeNetNotInterpretableETS, casal2023,herbold2023}.

Institutions have adopted proctoring tools such as video monitoring and browser lockdown systems to address academic dishonesty in online assessments \cite{tiong2021,roa2022}. However, these methods come with various limitations \cite{alsabhan2023,kamalov2021}. Screen lockdowns can be bypassed using secondary devices, while video surveillance requires costly human oversight and is invasive, raising privacy concerns  \cite{tiong2021,roa2022}.

In contrast, keystroke dynamics offers a low-intrusion, scalable alternative for detecting LLM-assisted cheating \cite{kundu2024,crossley2024,plank2016}. By capturing typing patterns, such as timing between key presses, keystroke analysis provides deeper insight into how the content was generated \cite{baaijen2012,zhang2021}. This helps to detect LLM-assisted cheating by revealing discrepancies in the underlying sub-processes, such as idea generation and revision patterns, between bona fide and assisted writings. \cite{flower1981,hayes2012,zhang2021}. 
 
While keystroke dynamics have been recently applied to detect academic dishonesty, existing research faces three key limitations. First, these studies are confined to English-language contexts, with a notable absence of publicly available keystroke datasets in languages other than English. This linguistic constraint hinders the generalization and applicability of keystroke-based cheating detection systems, especially in non-English environments \cite{plank2016, baaijen2012}. Second, current approaches focus primarily on binary classification, labeling writings as either bona fide or LLM-assisted \cite{casal2023, hoq2024}. This framing overlooks more subtle forms of cheating, such as paraphrasing or transcribing text generated by LLMs. Third, existing works overlook the role of cognitive load during writing, despite evidence that cognitive demands systematically influence keystroke patterns \cite{sellstone2023, balagani2013, vrij2006, zhang2021, hayes2012, flower1981}. Overlooking this variation limits a model's ability to account for task complexity. Therefore, we address these limitations by: 

\begin{enumerate}
    \item Constructing the first Korean-language keystroke dataset with $69$ users in the LLM-assisted cheating context, thereby expanding the linguistic and cultural scope of keystroke dynamics research.
    
    \item Introducing finer-grained classification of LLM-assisted writing into ChatGPT-response paraphrasing and transcribing, to better reflect the LLM-assisted cheating and evaluate the effectiveness of keystroke dynamics in detecting these behaviors.
    
    \item Systematically collecting keystroke data across six cognitive processes, as outlined in Bloom's Taxonomy (Table \ref{tab:cognitive_load_examples}), and analyzing the impact of cognitive awareness on the detection models ~\cite{anderson2001}.
\end{enumerate}



 
\section{Related Work}
\label{sec:related_work}
Keystroke dynamics—a measurement of timing patterns in typing behavior—have been widely studied in applications such as user authentication and identification \cite{joyce1990, leggett1988, fabianmonrose2000, kumar2016, kumar2018, KeystrokeVideo, KeystrokeSound}, authorship attribution, fake profile detection ~\cite{kuruvilla2024}, and inference of soft biometrics \cite{udandarao2020}. With the rise of LLM-powered tools like ChatGPT and their misuse in academic settings, keystroke analysis has gained renewed attention as a method for detecting LLM-assisted cheating and plagiarized writing~\cite{kundu2024, crossley2024}.

Kundu et al. \cite{kundu2024} utilized keystrokes to distinguish between bona fide and LLM-assisted writing. They created a dataset where students wrote in both unaided and ChatGPT-assisted conditions. Keystrokes were modeled and classified into either \textit{bona fide} or \textit{assisted} using a modified TypeNet architecture \cite{acien2022b}. The authors evaluated multiple configurations, including user-specific versus user-agnostic, keyboard-specific versus keyboard-agnostic, and context-specific versus context-agnostic settings. While their model achieved strong performance in controlled conditions (up to 85.7\% accuracy), accuracy declined in generalization settings, such as cross-user and cross-dataset evaluations. Their findings demonstrate the potential of keystroke dynamics for detecting LLM-assisted writing, while also highlighting key challenges in building models that generalize across diverse users, devices, and cheating scenarios.

Crossley et al.~\cite{crossley2024} addressed a related but distinct problem: detecting transcription of high-quality essays in controlled writing environments. Although they did not explicitly test LLM-assisted cheating, they note that such transcriptions could plausibly originate from generative models. Using a random forest classifier trained on keystroke features (e.g., \textit{pauses, insertions, deletions}), they achieved $99$\% accuracy in distinguishing transcribed argumentative essays from authentic ones. Their findings suggest that authentic writing involves \textit{cognitively complex behaviors}, such as planning, editing, and revision, which are observable in keystroke traces. In contrast, transcribed writing tends to be more linear and fluent \cite{sellstone2023, hayes2012}. Sellstone \cite{sellstone2023} further explored keystroke logging to analyze cognitive processes in a morphological knowledge test. Their results showed that pause and revision patterns—especially at morpheme boundaries—can capture real-time linguistic processing beyond what is reflected in final test scores.

Hart et al.~\cite{hart2023} investigated the deterrent effect of keystroke logging in a CS1 course, where students were required to submit keystroke logs along with their code. By comparing groups with and without logging, the authors found that logging reduced the self-reported temptation to plagiarize and slightly lowered the actual rates of plagiarism. While most students viewed the system positively, some reported increased anxiety and concerns about privacy. These results suggest that keystroke logging can serve as a low-overhead deterrent, though clear communication is essential to reduce stress and prevent misconceptions.

Zhang et al.~\cite{zhang2021} analyzed keystroke logs to infer cognitive processes during essay writing, using Hayes' model as a theoretical foundation. They segmented inter-key intervals into four states—Long Pause (LP), Text Production (TP), Local Editing (LE), and Global Editing (GE)—to approximate subprocesses like planning, translating, and revising. By dividing each writing session into ten equal-time segments, they captured the longitudinal distribution of these states and applied hierarchical clustering. This revealed three distinct writer profiles: linear writers (dominated by fluent text production), recursive writers (who alternated among planning, writing, and editing), and struggling writers (who showed delays and early editing). The study demonstrated that \textit{keystroke-based state transitions} reflect underlying \textit{cognitive strategies and time management}, and found that recursive writers tended to engage more deeply across sub-processes. However, differences in writing quality were modest.

To summarize, keystroke dynamics capture rich, process-level \textit{behavioral signals} that are \textit{inherently difficult to fake}, providing a strong foundation for detecting LLM-assisted cheating in writing. However, current approaches face several limitations. Most detection models are trained exclusively on English data~\cite{plank2016, kundu2024}, limiting their applicability in multilingual settings. Although keystroke datasets exist in several non-English languages—such as Chinese~\cite{Multi-Keyboard-Bilingual}, Arabic~\cite{ArabicKeystrokes}, Japanese~\cite{Japanese2009}, French~\cite{French2017}, Italian~\cite{ItalianKeystrokes2011}, Romanian~\cite{RomanianKeystrokes}, Russian~\cite{RussianKeystrokes}, and Korean~\cite{KoreanKeystrokes2020}—none were collected for evaluating LLM-assisted cheating detection. Additionally, LLM-assisted cheating is typically framed as a \textit{binary classification}—human versus LLM—overlooking more realistic scenarios such as paraphrasing, collaborative authorship, or LLM-assisted drafting~\cite{casal2023}. Moreover, few studies explicitly account for cognitive load, despite strong evidence that task complexity shapes writing behavior~\cite{sellstone2023, flower1981, zhang2021, hayes2012, balagani2013}.

To address these gaps and explore the full potential of keystroke dynamics, we present the first keystroke dataset for Korean writing tasks spanning six Bloom's Taxonomy-defined cognitive processes~\cite{anderson2001}. Our work distinguishes between bona fide writing, ChatGPT-response paraphrasing, and ChatGPT-response transcribing, and offers insights into keystroke-based detection of LLM-assisted cheating across varied cognitive contexts.

\begin{table*}[htp]
\centering
\caption{Carefully crafted questions to initiate six different levels of cognitive load via processes defined in Bloom's taxonomy \cite{anderson2001}.}
\label{tab:cognitive_load_examples}
\resizebox{0.99\textwidth}{!}{%
\begin{tabular}{ll}
\toprule
\textbf{Cognitive Processes} & \textbf{Example Questions Used} \\
\midrule
Remember   & Name two Korean historical figures and explain their significance. \\
Understand & How would you compare the role of women in traditional Korean society with their role in contemporary Korea? \\
Apply     & What should South Korea do to preserve its tradition and culture amidst rapid globalization? \\
Analyze     & Discuss the role of digital gaming culture in South Korea and its effects on youth socialization and education. \\
Evaluate    & Evaluate the extensive use of English loanwords in Korea. \\
Create      & How would you improve the inclusivity and accessibility of public spaces for people with disabilities in South Korea? \\
\bottomrule
\end{tabular}
} 
\vspace*{-11pt}
\end{table*}

\begin{figure}[htp]
    \centering
    \begin{subfigure}[t]{0.43\linewidth}
        \centering
    \includegraphics[width=2.22in, height=2.33in, keepaspectratio]{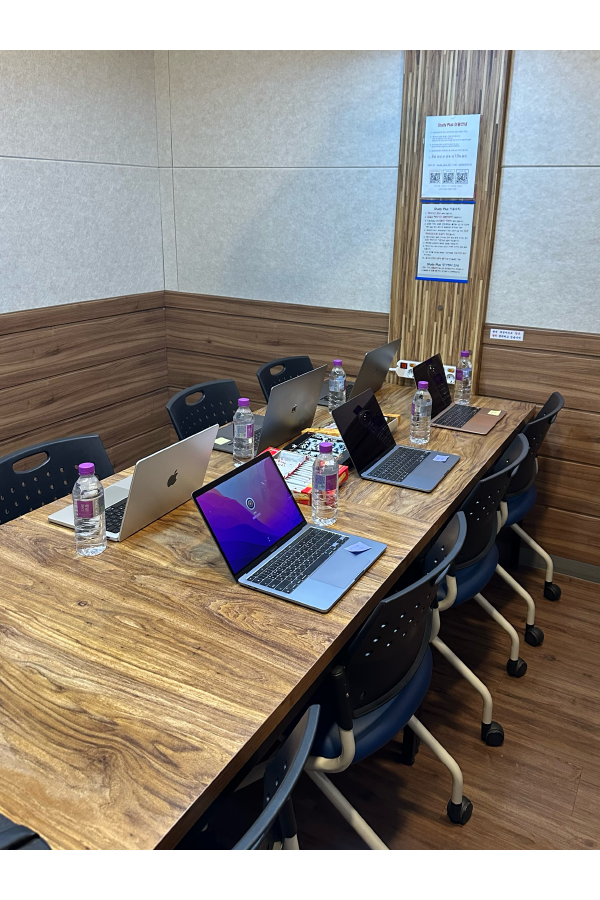}
        \caption{}
        \label{fig:data-collection-env}
    \end{subfigure}%
    \hspace{0.09in}
    \begin{subfigure}[t]{0.52\linewidth}
        \centering
        \includegraphics[trim=0 20 0 0, width=\linewidth]{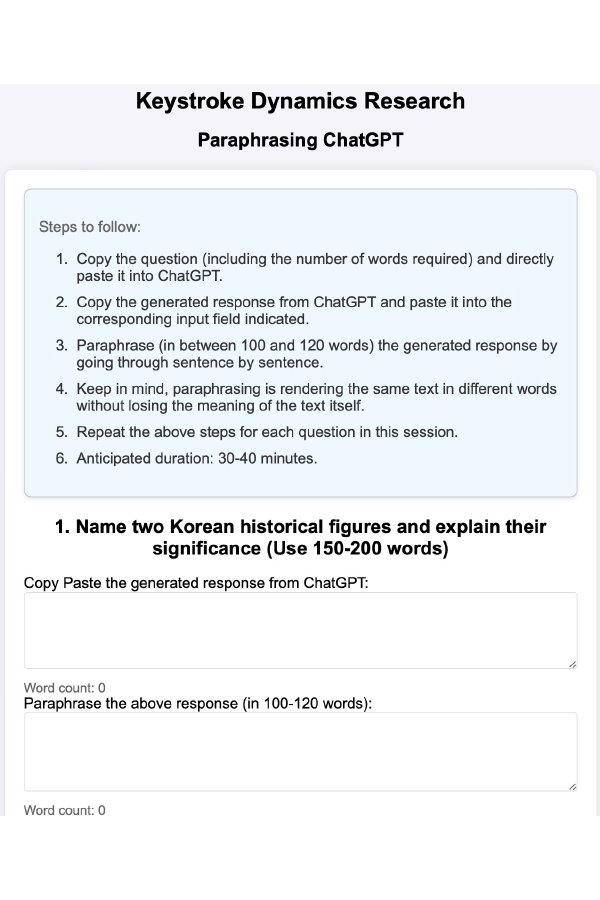}
        \caption{}
        \label{fig:data-collection-website}
    \end{subfigure}
    \caption{Overview of the data collection setup. Figure \ref{fig:data-collection-env} shows the data collection environment. Figure \ref{fig:data-collection-website} shows a snapshot of the web portal we created. The portal is easily customizable for other languages or question sets.}

    \label{fig:data-collection} \vspace{-0in}
\end{figure}



\section{Material and Methods}
\label{sec:materials_and_methods}
\subsection{Design of data collection experiment}
The unique objectives of this study necessitated the collection of Korean keystroke data in the context of academic dishonesty. Thus, we first obtained approval from the Institutional Review Board (IRB) for the study and data collection. 

We created a data collection website to capture keystroke data, including KeyType, KeyEvent (Down/Up), and Timestamp, across the three writing scenarios.
\underline{Bona fide writing}: independently generating original responses without using ChatGPT or any other external assistance.
\underline{ChatGPT-response paraphrasing}: rephrasing ChatGPT-generated responses using synonyms and altering sentence structures while retaining the original meaning.
\underline{ChatGPT-response transcribing}: directly retyping ChatGPT-generated responses, reproducing the content word-for-word.

For each writing scenario, we included six prompts, each corresponding to one of the six cognitive processes outlined in Bloom's Taxonomy \cite{anderson2001}. As illustrated in Table \ref{tab:cognitive_load_examples}, the prompts were crafted to avoid requiring domain-specific knowledge while focusing on general, culturally relevant topics for Korean university students, who were the primary target participants given the study's focus on academic dishonesty in the Korean language. This design choice helped isolate the potential confounding effects of topic familiarity, allowing for a clearer interpretation of the findings.  

We provided an input field and a real-time word count tracker for each question in every typing session. In the paraphrasing and transcribing sessions, a secondary input field was included for each question, allowing the participants to paste the ChatGPT-generated response they obtained and refer to it while paraphrasing or transcribing based on the current writing scenario. This setup allowed for collecting both the ChatGPT-generated text and the participant-crafted response, enabling comparison. This comparison helped validate whether participants adhered to the assigned writing condition and ensured the integrity of the collected data. To further maintain task integrity, we \textit{disabled copy-pasting} in all input fields except those secondary input fields designated for pasting ChatGPT-generated responses.

Moreover, the required word count for each response across the three writing scenarios was carefully determined by examining the time taken by the initial two participants. In particular, we reduced the word count from $200$ to $100$–$120$ words, and introduced a five-minute break between the three writing scenarios. This adjustment brought the total expected data collection duration to approximately $2$ hours for each participant. Also, it ensured that the prolonged typing did not distort natural typing behavior due to fatigue or loss of focus.

Additionally, preliminary testing revealed that ChatGPT frequently produced shorter outputs when prompted in Korean to generate responses within the $100$–$120$ word range. To mitigate this inconsistency, the word count range specified at the end of each question, used as the input prompt for ChatGPT, was slightly increased in paraphrasing and transcribing sessions. Participants were informed in advance that there was a discrepancy between the word count stated in the prompt and the target range for their responses in those sessions, to prevent confusion. 

\subsection{Data collection}
We recruited $69$ university students through an advertisement posted on a university community platform in Korea. The sample included $33$ females and $36$ males, with a mean age of approximately $23$ years. Before participating, all individuals received a consent form that outlined the study’s objectives, procedures, and potential risks. Data collection was conducted in person to maintain control over the experimental conditions and ensure the quality of the data.

The experiment consisted of two phases. The second phase replicated the structure of the first but used different question sets to minimize potential carry-over effects. Of the $69$ participants, $50$ completed both phases. For this paper, we used only the first-phase data to emphasize user-independent evaluation. The second-phase data will be used in future work for user-dependent evaluation, following the methodology described in \cite{kundu2024}. 

We provided on-site support to address technical issues, clarify instructions, and ensure participants adhered to the multi-faceted writing task design. This oversight was important given the complexity of the data collection methodology, which aimed to capture nuanced differences in LLM-assisted writings and varying cognitive load levels. Thus, we employed three validation steps during data collection: $(1)$ In the paraphrasing and transcribing sessions, participants were instructed to generate responses to each prompt using ChatGPT and notify the researcher before proceeding. The researcher verified the word count and, if necessary, assisted with re-prompting to obtain a response within the appropriate word count range. This ensured consistency and uniform task conditions across participants. $(2)$ Upon completion of one writing scenario, participants were required to notify the researcher before proceeding to the next. Their responses were then reviewed to confirm appropriate task engagement, adequate word count, and adherence to scenario-specific instructions. $(3)$ Besides post-session validation, we provided sufficient examples of paraphrasing and periodically checked to ensure participants met the minimum expected level of paraphrasing. This reduced the possibility of paraphrasing keystroke patterns to resemble those of transcribing the ChatGPT responses.  

Participants also provided demographic information, including age, gender, and handedness. In addition to keystroke data, we saved user and ChatGPT-generated text responses to conduct text and behavior-level analysis. This dataset is also suitable for studying Keystroke-based authentication and authorship attribution in the Korean language. The dataset and code are publicly available\footnote{\url{https://github.com/rajeshjnu2006/kkcd-ijcb2025}}. 

\subsection{Data preprocessing}
Raw keystroke data were preprocessed to correct inconsistencies in key event sequences. We first removed rare but systematic errors where the key was logged as "Unidentified," which consistently followed a CapsLock Down event—used by participants to switch input languages when typing English loanwords.

Prolonged key presses (e.g., holding Backspace or Arrow keys to delete or navigate) were logged as repeated KeyDown events followed by a single KeyUp. We retained only the first KeyDown and the final KeyUp in each sequence to ensure accurate extraction of hold and interval times.

Furthermore, we corrected misclassifications between base characters and their Shift-modified counterparts by analyzing key event context within Shift sequences (Table~\ref{tab:keystroke_with_correction}). These errors often arose during rapid typing when a participant pressed a base character (e.g., a consonant or vowel) and activated the Shift key before releasing the base character key. This caused the system to mislabel the KeyUp event as the Shift-modified variant. For instance, as shown in Table \ref{tab:keystroke_with_correction}, at time $t_3$, the Korean base alphabet "ㄱ" was incorrectly logged as the Shift-modified version "ㄲ" due to the Shift key being active. Conversely, some Shift-modified characters were misrecorded as base forms when the Shift key was released too early—e.g., at $t_6$, "ㅖ" was logged as "ㅔ". All corrections were logged for traceability, and a final validation step ensured the completeness and consistency of key event pairs.

\begin{table}[htp]
\centering
\small
\caption{Keystroke sequence with raw and corrected sequence.}
\label{tab:keystroke_with_correction}
\renewcommand{\arraystretch}{1.15}
\begin{tabular}{p{1cm} p{1.3cm} p{1.2cm} p{0.9cm} p{1.1cm}}
\toprule
\textbf{Raw} & \textbf{Corrected} & \textbf{Code} & \textbf{Event} & \textbf{Time} \\
\midrule
\texttt{ㄱ}    & \texttt{ㄱ}    & KeyR       & Down & $t_1$ \\
Shift         & Shift          & ShiftLeft  & Down & $t_2$ \\
\texttt{ㄲ}    & \texttt{ㄱ}    & KeyR       & Up   & $t_3$ \\
\texttt{ㅃ}    & \texttt{ㅃ}    & KeyO       & Down & $t_4$ \\
Shift         & Shift          & ShiftLeft  & Up   & $t_5$ \\
\texttt{ㅂ}    & \texttt{ㅃ}    & KeyO       & Up   & $t_6$ \\
\bottomrule
\end{tabular}
\vspace*{-10pt}
\end{table}

\subsection{Keystroke modeling}
To capture both low-level motor behavior and higher-order cognitive processes, we extracted two feature sets from the preprocessed keystroke data: \textit{Temporal features} and \textit{Rhythmic features}. Temporal features include Key Hold Times (KHT) and Key Interval Time (KIT), which reflect typing fluency and physical execution. Rhythmic features capture cognitively driven behaviors such as planning, revision, and structural pausing, as grounded in cognitive writing models.

While recent work has explored end-to-end approaches \cite{Stragapede2023KVC}, including models like TypeNet~\cite{acien2022b, kundu2024} and TypeFormer \cite{stragapede2024typeformer}, we focus primarily on \textit{interpretable feature-based modeling}. In addition, prior studies have shown that handcrafted Temporal features can match or even surpass deep models in performance~\cite{wahab2023}, and Rhythmic features \cite{balagani2013,crawford2015} offer transparency critical for analyzing cognitive effort and LLM involvement. In contrast, TypeNet architectures often suffer from limited interpretability~\cite{TypeNetNotInterpretableETS}. Developing scalable yet interpretable architectures for keystroke modeling remains an open direction for future work. In addition, it would be interesting to explore a hybrid model that uses both content (response text) and behavior (keystroke dynamics) \cite{khurana2023}.

\subsubsection{Temporal feature extraction}
After preprocessing, we extracted two standard keystroke features per user: KHT and KIT from the keystroke timestamps~\cite{kuruvilla2024,acien2022,shadman2023}. Backspace and Arrow keys occasionally yielded implausibly short durations ($0-2$ms), likely due to rapid key repetition; these outliers were excluded. For KIT, intervals were computed only between consecutive KeyDown events within the same question to maintain contextual consistency. 

We consolidated all users' processed KHT and KIT data into a structured database indexed by user ID and task identifier. Each identifier followed the format $x.y$—with $x$ denoting the writing scenario ($0$: bona fide, $1$: paraphrasing, $2$: transcribing) and $y$ the question number $(1$–$6)$. This structure enabled flexible access to data partitions for Cognition-Aware and Cognition-Unaware analyses.

To construct the Temporal feature matrix, we selected a common KHT and KIT feature set based on the observed key and key pairs for each user. We computed five summary statistics for each feature—first quartile, median, third quartile, mean, and standard deviation—yielding fixed-length feature vectors. Outlier filtering was applied at the feature level using the 0.5th and 99.5th percentiles to ensure robustness against noise. Missing values were imputed using feature-wise medians.

\subsubsection{Rhythmic feature extraction}

Rhythm-based features such as pauses, bursts, deletions, and insertions are effective proxies for core cognitive sub-processes of writing—planning, evaluation, transcription, and revision—as outlined in cognitive writing models~\cite{hayes2012}. Derived directly from keystroke data, these features provide interpretable behavioral signals related to cognitive load~\cite{crossley2024, sellstone2023, wengelin2006, balagani2013, banerjee2014}, and have been applied to detect LLM-assisted writing across varied cognitive contexts~\cite{trezise2019, crossley2024}.

While Crossley et al.~\cite{crossley2024} employed fixed pause thresholds ($200$ms and $2000$ms) to distinguish motor actions from higher-order processing, we adopted a binned approach using $300$ms intervals from $(3$ms$,300$ms$)$ to $(2400$ms$+)$ to capture finer-grained pause dynamics and mitigate potential mimicry. Additionally, we extracted inter-word, inter-sentence, and pre-deletion pauses to model lexical, syntactic, and evaluative planning.

For burst features, we included: \textit{P-burst}, defined as continuous text production segments terminated by a pause $\ge 2$ seconds; \textit{R-burst}, a modification of the revision burst~\cite{crossley2024}, capturing writing sequences ending in deletion; and \textit{Delete burst}, representing consecutive deletion events.

\textit{Aggregate fluency metrics (e.g., characters, words, or sentences per minute) were excluded because they are susceptible to manipulation through artificial pacing.}

We extracted \textit{fifteen} rhythm-based feature types: nine binned pauses, P-burst, R-burst, delete burst, inter-word pause, inter-sentence pause, and pause-before-delete. Each was a list of values, so we computed seven statistics—\textit{mean, standard deviation, total duration, count, median, and first and third quartiles}. We also added entropy for pause durations and P-burst lengths, resulting in a total of $107$ features.  

Among the most discriminative features were long pauses (e.g., $>2400$ms), P-burst statistics (mean and entropy), and inter-sentence pauses—reflecting cognitive load and planning effort. Revision-sensitive features such as R-bursts and delete bursts also contributed meaningfully. While short-to mid-range pauses may be consciously manipulated, revision-related behaviors and entropy-based measures are less susceptible to disguise.  

\subsection{Experimental scenarios}

We employed a user-independent setup, where training and testing were performed on non-overlapping subsets of the 69 participants to evaluate generalization to unseen users and prevent overfitting to individual typing patterns. To assess the effect of cognitive load, we defined two conditions: \textit{Cognition-Unaware} and \textit{Cognition-Aware}.

In the \textit{Cognition-Unaware} condition, keystroke data from all prompts were aggregated per user, ignoring differences in cognitive load. This served as a baseline for evaluating overall model performance.

In the \textit{Cognition-Aware} condition, prompts were split by cognitive processes and their levels in Bloom's Taxonomy: questions $1$–$3$ as low (L) and $4$–$6$ as high (H) load (see Table \ref{tab:cognitive_load_examples}). We evaluated four configurations: H$\rightarrow$H, H$\rightarrow$L, L$\rightarrow$H, and L$\rightarrow$L, corresponding to train$\rightarrow$test combinations. This setup enabled the evaluation of the model's sensitivity to cognitive variability.

\subsection{Training and testing splits}
We generated $105$ unique train-test splits by varying the training set size from $30$\% to $70$\% in $2$\% increments, with $5$ random trials at each training percentage to ensure robust generalization. For each split, a subset of users (from the pool of users $1-69$) was randomly selected for training based on the specified percentage, and the remaining users were assigned to the test set, ensuring a user-independent evaluation. The same train-test splits were applied consistently across both the \textit{Cognition-Unaware} and \textit{Cognition-Aware} scenarios.

\begin{figure*}[htp]
    \centering
    \includegraphics[width=5.7in, height=1.6in, trim=0.1in 0in 0in 0in, clip]{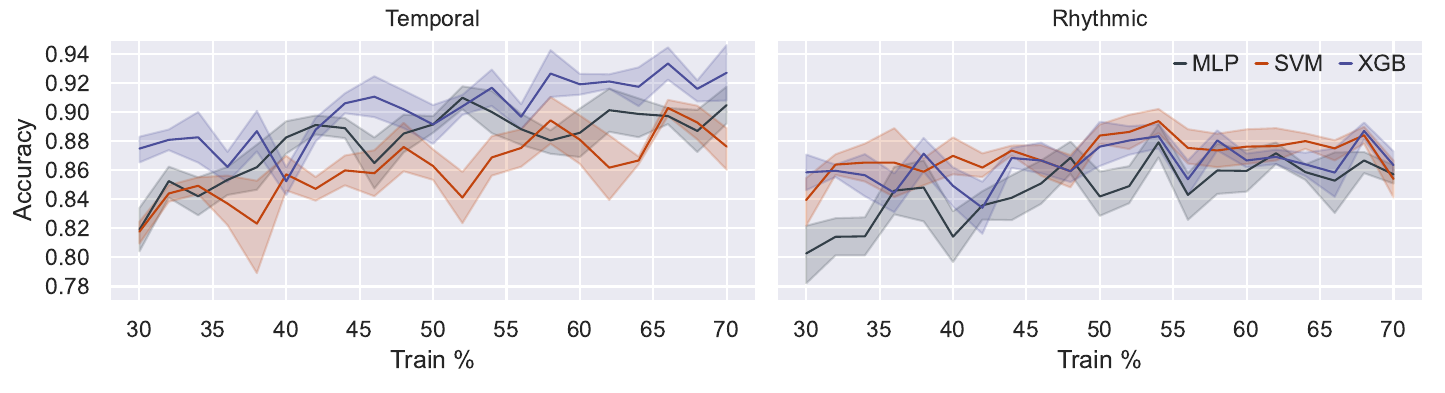}
    \caption{Performance of Temporal and Rhythmic feature-based models under Cognition-Unaware conditions.}
    \label{fig:unaware-combined}
    \vspace*{-4pt}
\end{figure*}

\begin{figure*}[htp]
    \centering
    \includegraphics[width=6.66in, height=1.11in]{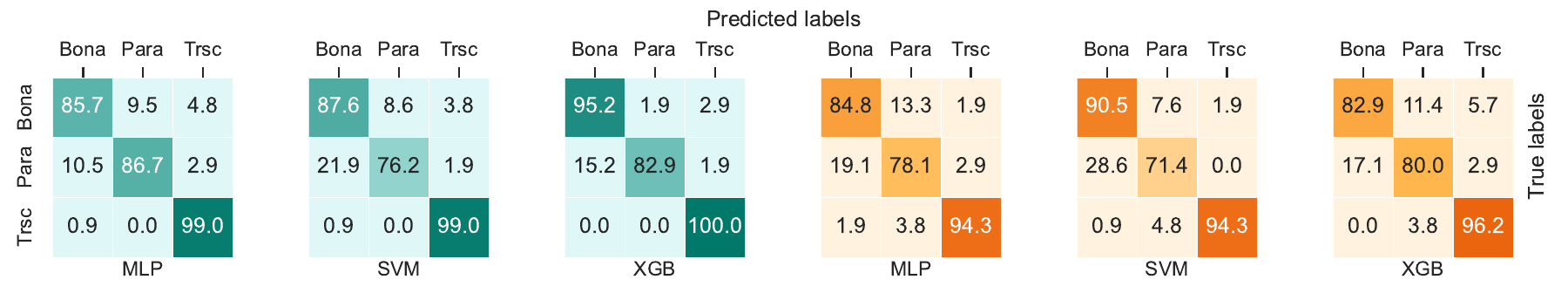}
    \caption{Class-wise performance under Cognition-Unaware conditions. Temporal features in Teal blue, and Rhythmic features in Orange.}
    \label{fig:unaware-combined-cm}
\end{figure*}
 
\subsection{Classifiers and hyperparameter tuning}
We evaluated three classifiers: Multi-Layer Perceptron (MLP), Support Vector Machine (SVM), and Extreme Gradient Boosting (XGB). These models were selected based on their superior performance in prior keystroke classification~\cite{crossley2024, TypeNetNotInterpretableETS, MLClassKeystrokeFairness}. We used mutual information for each train–test split to select the top $50\%$ of features from the training set, and applied the same subset to the test set. Hyperparameters were optimized using a Genetic algorithm \cite{young2015optimizing}. Each candidate solution encoded a unique configuration and was evaluated using 5-fold stratified cross-validation with accuracy as the fitness metric. The algorithm refined the population over generations through crossover, mutation, and selection. The best configuration was used to train the final model, which was evaluated on the test set. 

\subsection{Evaluation metrics} We evaluate model performance using accuracy, recall, and the confusion matrix to show classification performance for each class. Accuracy measures the overall proportion of correctly classified samples across all classes. For each class, recall measures the proportion of actual instances that are correctly identified (e.g., of all truly transcribed samples, how many were correctly predicted as transcribed). The confusion matrix provides a comprehensive breakdown of predictions versus ground truth across all class pairs, offering insight into common misclassification patterns, such as paraphrased samples being mislabeled as bona fide or vice versa. 

To compute accuracy, we average the accuracy from five trials for each train-test split. Additionally, rather than using all $105$ confusion matrices, we aggregated the five confusion matrices from the standard $70-30$ splits to generate the final confusion matrix for each model.

\section{Results}
We report four sets of results, cognition-aware, cognition-unaware, cross-cognition, and human baseline. Models consistently performed best on detecting transcribed responses, with paraphrased responses showing the most variability across conditions.

\subsection{Cognition-Unaware}
Figure~\ref{fig:unaware-combined} shows accuracy trends of Temporal (left) and Rhythmic (right) models across increasing training percentages under Cognition-Unaware conditions. 

For Temporal models, XGBoost consistently outperformed MLP and SVM across most training sizes. Its accuracy ranged from approximately $87$\% at $30$\% training data to around $93$\% at $70$\% training data. MLP showed steady improvement as training data increased, with accuracy rising from $83$\% to about $90$\%. SVM trailed behind between ($30-50$\%) training data but narrowed the gap beyond $60$\% training data, stabilizing close to $89$\% accuracy. Temporal models also exhibited reduced variance as the training size increased, with tighter confidence bands, particularly for XGBoost, indicating more stable learning.

In contrast, Rhythmic models exhibited more modest accuracy gains. All three classifiers plateaued around 40\% training data and displayed minor improvements with additional training data. XGBoost began at $86$\% with $30$\% training data, fluctuating between $85-88$\% through mid-range splits, and peaked just below $89$\% near $70$\% training data. SVM achieved similar accuracy with a slightly smoother curve, while MLP trailed behind in earlier splits ($\approx 80$\% accuracy at $30$\% training data) but converged toward $87-88$\% with more data. The Rhythmic models showed less performance spread across classifiers, with accuracy curves overlapping more closely than in the Temporal case. This suggests that Rhythmic features offer more constrained discriminative capacity, and increasing data does not boost performance as significantly as with Temporal features.

Overall, Temporal features enabled stronger and more scalable generalization as training size increased, particularly for XGBoost. While stable, Rhythmic models showed limited sensitivity to training size, reinforcing earlier findings that they capture less nuanced behavioral signals than Temporal keystroke dynamics.

Figure~\ref{fig:unaware-combined-cm} shows class-specific confusion matrices for Temporal (left, teal) and Rhythmic (right, orange) models under Cognition-Unaware conditions using a $70$-$30$ train-test split.

The \uline{Transcribed} class was consistently the most reliably detected across all models. Temporal XGBoost achieved perfect recall ($100$\%), while MLP and SVM followed closely at $99.05$\%. Rhythmic models showed a modest drop, with XGBoost at $96.19$\%, and both MLP and SVM at $94.29$\%. These results suggest that transcribed responses, which tend to follow highly fluent and linear keystroke patterns, are readily distinguishable, even with limited temporal resolution.

\begin{figure*}[htp]
\centering
\includegraphics[width=6.3in, height=3.6in]{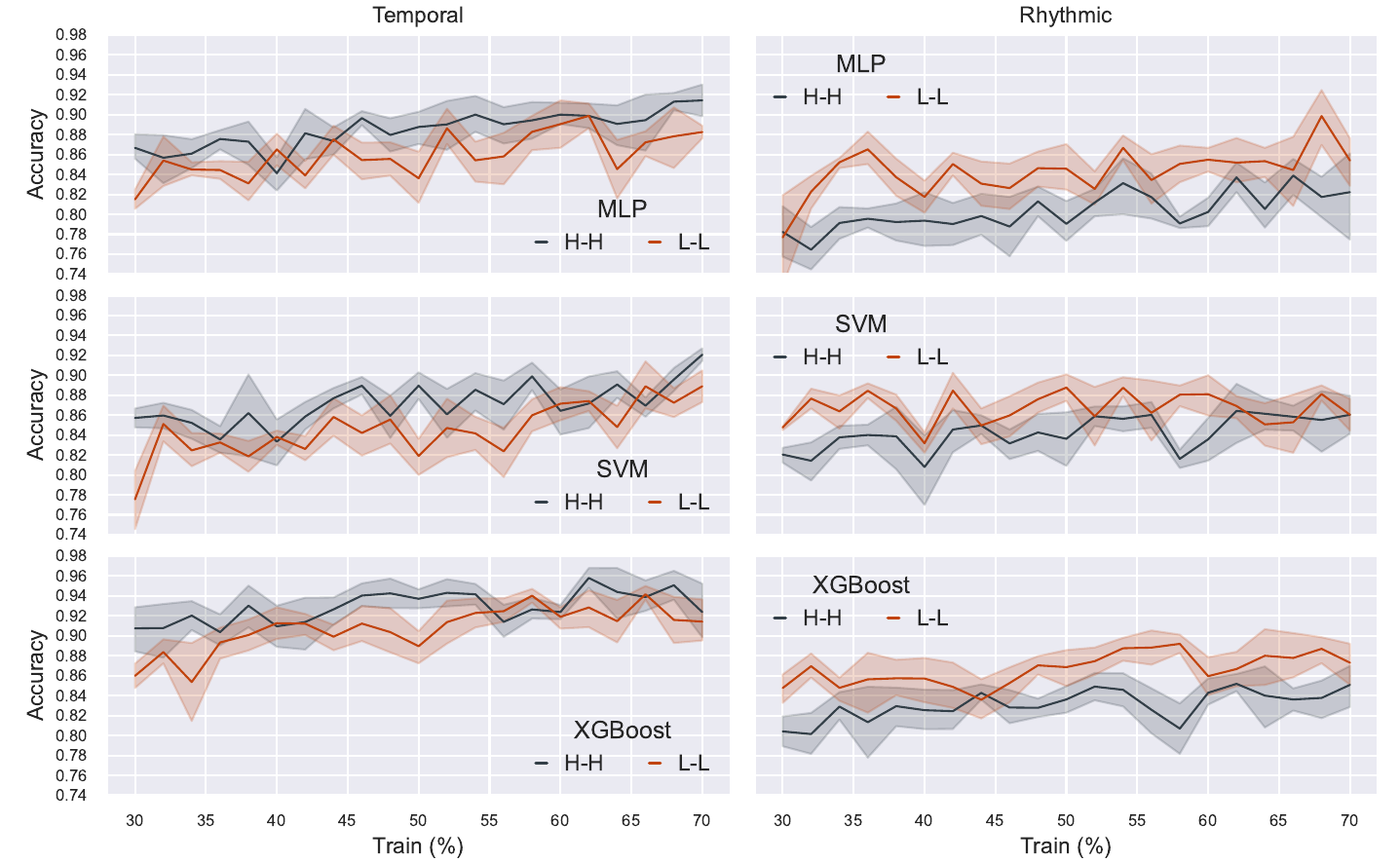}
    \caption{Performance of Temporal and Rhythmic feature-based models under Cognition-Aware (H$\rightarrow$H and L$\rightarrow$L) conditions over different train-test splits.}
    \label{fig:aware-combined}
\end{figure*}

\begin{figure*}[htp]
    \centering
    \includegraphics[width=6.6in, height=2.2in]{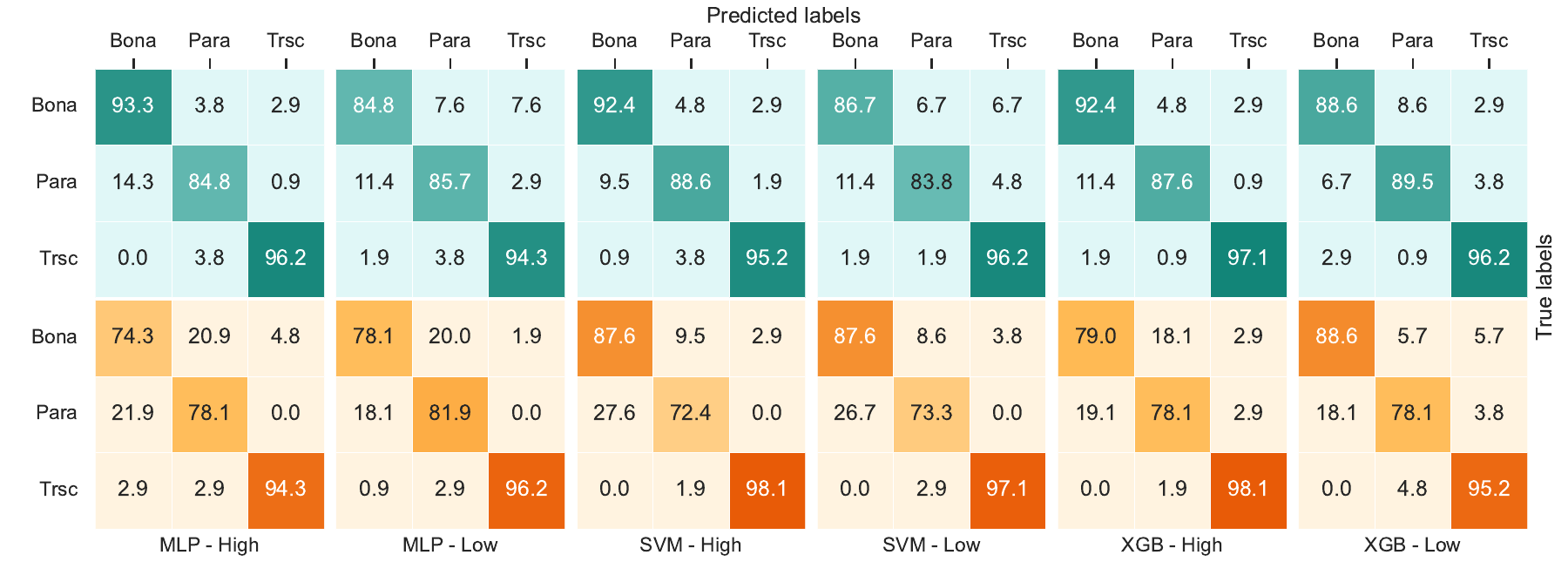}
    \caption{Class-wise performance under Cognition-Aware conditions. Temporal features in Teal blue, and Rhythmic features in Orange. High means H$\rightarrow$H, i.e., trained on High and tested on High. Likewise, Low means L$\rightarrow$L, i.e., trained on Low and tested on Low.}
    \label{fig:aware-combined-cm-both}
\end{figure*}

The \uline{Paraphrased} class showed the most variability. Temporal models performed best with MLP at $86.67$\% recall, followed by XGBoost at $82.86$\% and SVM at $76.19$\%. Rhythmic performance was weaker overall: XGBoost achieved $80.0$0\%, MLP $78.10$\%, and SVM dropped to $71.43$\%. Notably, SVM and MLP showed increased confusion between paraphrased and bona fide responses, indicating that Rhythmic features may struggle to capture the subtle planning and revision signals necessary to differentiate paraphrased writing from authentic writing.

For the \uline{Bona fide} class, XGBoost again led among Temporal models with 95.24\% recall, while SVM and MLP achieved $87.6$2\% and $85.71$\%, respectively. Rhythmic performance dropped slightly for SVM ($90.48$\%) and MLP ($84.76$\%), and more noticeably for XGBoost ($82.86$\%). The consistent drop across all Rhythmic models was primarily due to misclassification into the paraphrased class, underscoring the importance of temporal precision in identifying genuine, cognitively effortful writing.

In sum, Temporal features provided stronger and more consistent discriminative power under Cognition-Unaware conditions, particularly for separating paraphrased and bona fide responses. Rhythmic features remained robust for detecting transcribed text. Still, they showed reduced reliability for cognitively complex or stylistically subtle distinctions, highlighting the importance of fine-grained timing signals in keystroke-based LLM writing detection.

\subsection{Cognition-Aware}
Figure \ref{fig:aware-combined} suggests that Temporal feature-based models consistently show higher accuracy under the (H$\rightarrow$H) setting compared to (L$\rightarrow$L), with the margin widening as training size increases—most notably in XGBoost, which surpasses $93$\% accuracy beyond the $60$\% training mark. This indicates that high-load keystroke data offers more discriminative temporal patterns for model learning.

In contrast, Rhythmic models perform better under L$\rightarrow$L than H$\rightarrow$H across all classifiers, particularly for MLP and SVM, suggesting that Rhythmic features are more stable and learnable when cognitive load is low. With increasing training data, Rhythmic H$\rightarrow$H performance remains flat or noisy, highlighting weaker generalization under high-load variability. XGBoost maintains the most stable performance across feature types and cognitive loads, though Temporal H$\rightarrow$H remains its strongest regime. These patterns reveal a modality-load interaction: Temporal features benefit from high-load cognitive complexity, while Rhythmic features degrade in such settings. These complementary trends suggest that \textit{fusing Temporal and Rhythmic features may improve generalization}, as each modality captures distinct behavioral signals and can compensate for the other's weaknesses under varying cognitive loads. \textit{We plan to investigate this in the future.} 

Figure~\ref{fig:aware-combined-cm-both} presents the class-specific confusion matrices for Temporal (top, teal) and Rhythmic (bottom, orange) models under Cognition-Aware conditions, evaluated under H$\rightarrow$H and L$\rightarrow$L scenarios using a $70-30$ split.

The \uline{Transcribed} class remained the most reliably detected across all models. Temporal XGBoost achieved the highest recall ($97.14$\% under high load), followed by SVM ($95.24$\% high, $96.19$\% low) and MLP ($96.19$\% high, $94.29$\% low). Rhythmic models also performed strongly in this class, with XGBoost reaching $98.10$\% (high) and $95.24$\% (low), and all other models exceeding $94$\% in both conditions. This suggests that both feature types robustly capture surface-level fluency patterns typical of transcribing.

The \uline{Paraphrased} class showed greater variation. Temporal models performed best with XGBoost ($87.62$\% high, $89.52$\% low), followed by SVM ($88.57$\% high, $83.81$\% low), and MLP ($84.76$\% high, $85.71$\% low). Rhythmic models struggled more—SVM dropped to $72.38$\% (high) and $73.33$\% (low), and MLP showed modest improvement from $78.10$\% (high) to $81.90$\% (low). Across settings, misclassifications primarily occurred between paraphrased and bona fide, especially under high cognitive load where stylistic overlap intensifies.

The \uline{Bona fide} class was best handled by Temporal MLP ($93.33$\% high), followed by XGBoost and SVM (each $92.38$\% high). Under low load, recall declined slightly across models, with XGBoost at $88.57$\%, MLP at $84.76$\%, and SVM at $86.67$\%. Rhythmic models showed larger drops—MLP reached only $74.29$\% (high), and XGBoost $79.05$\%, with common confusion into paraphrased responses, indicating reduced separability without precise timing cues.

In summary, Temporal features provided more stable and accurate classification across cognitive loads, particularly in distinguishing between paraphrased and bona fide writing. Rhythmic features remained effective for transcribed detection but were less robust under increased cognitive demand, particularly in distinguishing closely related writing types, such as paraphrased and authentic text.

\subsection{Cross-cognition} 
\begin{figure}
    \centering
    \includegraphics[width=3.24in, height=1.25in]{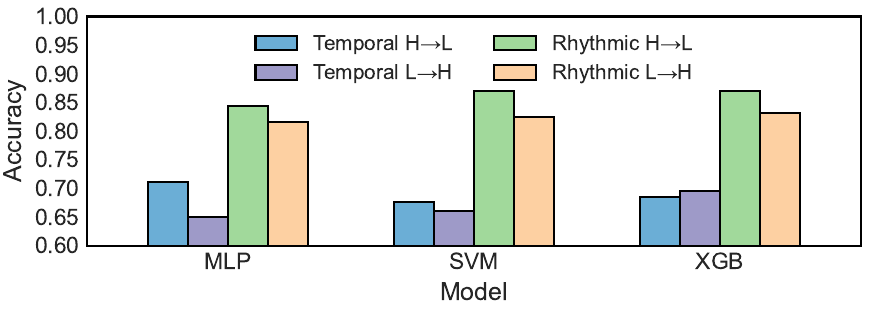}
    \caption{Performance of Temporal and Rhythmic feature-based models in cross-cognition environments.}
    \label{fig:cross_cog_combined} 
    \vspace{-0.1in}
\end{figure}

Figure~\ref{fig:cross_cog_combined} compares model accuracy under cross-cognition scenarios. Temporal models showed poor generalization, with accuracy dropping below $72$\% in all cases, highlighting their sensitivity to mismatched cognitive load. In contrast, Rhythmic models consistently outperformed their Temporal counterparts, maintaining high accuracy ($81$–$87$\%) across both H$\rightarrow$L and L$\rightarrow$H transfers. The small performance gap between transfer directions in Rhythmic models suggests that Rhythmic features are more robust to shifts in cognitive demand and better suited for real-world settings with task variability.

\subsection{Human baseline} 
\label{human_baseline}
We conducted a web-based survey to evaluate human ability to detect LLM-assisted writing under a Cognition-Unaware setting. Eighteen fluent Korean-speaking participants—primarily students and academics—were shown $18$ randomly selected participant responses ($6$ Bona fide, $6$ Paraphrased, $6$ Transcribed) through a custom-built interface. For each response, participants were asked to choose one label—\textit{Bona fide}, \textit{Paraphrased}, or \textit{Transcribed}—without being provided labeled examples. However, minimal contextual guidance and explicit definitions of each label were provided. Bona fide responses were correctly identified $69.44\%$ of the time; paraphrased responses, $41.67\%$; and transcribed responses, $49.07\%$. Paraphrased texts were most frequently misclassified, often labeled as transcribed ($37.96\%$) or bona fide ($20.37\%$). These findings illustrate humans' challenges in discerning nuanced forms of LLM-assisted writing, underscoring the value of keystroke-based behavioral detection systems.

\section{Conclusion} We introduced the first keystroke-based framework for detecting LLM-assisted cheating in Korean, addressing critical gaps in language coverage, cognitive context, and the granularity of cheating behavior. Our dataset captures fine-grained typing patterns across Bona fide, Paraphrased, and Transcribed writing scenarios under six levels of cognitive load induced by processes listed in Bloom's Taxonomy. By extracting and evaluating Temporal and Rhythmic features under Cognition-Aware and -Unaware conditions, we demonstrated that Temporal features offer stronger discriminative power, particularly in high cognitive load settings. In contrast, Rhythmic features provide robustness across cognitive load variability.  

\balance
{\small
\bibliography{references}
}

\end{document}